\pdfoutput=1

\documentclass[11pt]{article}
\usepackage{enumitem}

\usepackage[final]{acl}

\usepackage{times}
\usepackage{latexsym}
\usepackage{graphicx}
\usepackage{adjustbox}
\usepackage{array}

\usepackage[T1]{fontenc}

\usepackage[utf8]{inputenc}

\usepackage{microtype}

\usepackage{inconsolata}

\usepackage{graphicx}

\title{Non-Contextual BERT or FastText? A Comparative Analysis }

\author{
    Abhay Shanbhag\textsuperscript{1,3}, Suramya Jadhav\textsuperscript{1,3}, Amogh Thakurdesai\textsuperscript{1,3}, 
    Ridhima Sinare\textsuperscript{1,3}, and Raviraj Joshi\textsuperscript{2,3} \\
    \textsuperscript{1}Pune Institute of Computer Technology, Pune \\
    \textsuperscript{2}Indian Institute of Technology Madras, Chennai \\
     \textsuperscript{3}L3Cube Labs, Pune
}

\begin{document}
\maketitle
\begin{abstract}
Natural Language Processing (NLP) for low-resource languages, which lack large annotated datasets, faces significant challenges due to limited high-quality data and linguistic resources. The selection of embeddings plays a critical role in achieving strong performance in NLP tasks. While contextual BERT embeddings require a full forward pass, non-contextual BERT embeddings rely only on table lookup. Existing research has primarily focused on contextual BERT embeddings, leaving non-contextual embeddings largely unexplored. In this study, we analyze the effectiveness of non-contextual embeddings from BERT models (MuRIL and MahaBERT) and FastText models (IndicFT and MahaFT) for tasks such as news classification, sentiment analysis, and hate speech detection in one such low-resource language—Marathi. We compare these embeddings with their contextual and compressed variants. Our findings indicate that non-contextual BERT embeddings extracted from the model's first embedding layer outperform FastText embeddings, presenting a promising alternative for low-resource NLP.

\end{abstract}

\section{Introduction}
Word embedding is a way of representing words into dense vectors in a continuous space such that the vectors capture the semantic relationship between the words for the models to understand the context and meaning of the text. FastText, a context-independent method, basically captures the subword information, enabling it to learn rare words, misspelled words, and out-of-vocabulary words. It is recognized in the NLP community for its efficient performance in tasks like text classification and sentiment analysis. Despite being relatively old, it still remains one of the most effective alternatives when performing tasks on large datasets across various languages due to its subword-based approach.

\begin{figure*}
    \centering
    \includegraphics[width=1\linewidth]{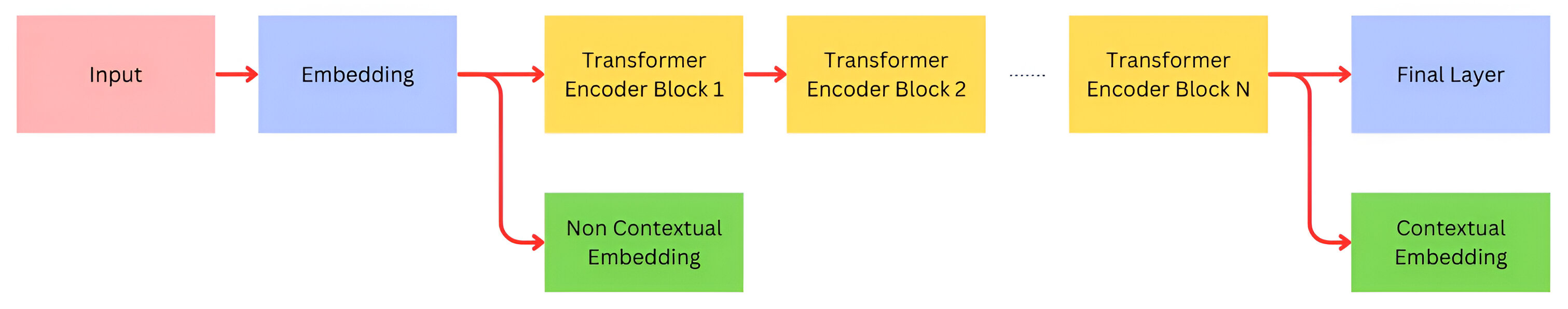}
    \caption{Embedding extraction workflow for contextual and non-contextual representations}
    \label{fig:BERT_Arch}
\end{figure*}

BERT (Bidirectional Encoder Representations from Transformers) \cite{DBLP:journals/corr/abs-1810-04805} word embeddings understand the meaning of a word based on its context in a sentence. The embeddings extracted just before the first embedding layer of the BERT architecture are referred to as non-contextual embeddings, while those obtained from the last hidden layer of BERT are known as contextual embeddings (Refer Fig \ref{fig:BERT_Arch}). Numerous variations of BERT like IndicBERT \cite{kakwani-etal-2020-indicnlpsuite}, MuRIL \cite{khanuja2021muril}, AfriBERT \cite{ralethe-2020-adaptation}, and mBERT \cite{DBLP:journals/corr/abs-1810-04805} to name a few, are available for experiments.

Recent studies have experimented with both FastText and BERT for various tasks; however, most of them focus on exploring contextual BERT embeddings. Experiments of \citet{DSa2020BERTAF} demonstrated that BERT embeddings outperformed FastText for classifying English text into toxic and non-toxic. Findings of \citet{Ahmed2024EnhancingDP} suggested that BERT embeddings outperformed those of FastText with an F1 score of 84\% when evaluated for depressive post-detection in Bangla.




While BERT consistently outperforms other word embeddings in various tasks for high-resource languages (HRLs) like English (\citet{Malik2021ToxicSD}), its effectiveness in low-resource languages (LRLs) remains relatively underexplored. This gap is particularly pronounced when balancing model performance with computational efficiency, which becomes a critical factor in low-resource settings.

Previous studies (\citet{DSa2020BERTAF}) have focused on contextual BERT embeddings, which outperform FastText due to their ability to capture contextual information. However, the use of non-contextual BERT embeddings for classification tasks in low-resource languages like Marathi remains unexplored. Unlike contextual embeddings, which require a full forward pass through the model, non-contextual embeddings can be obtained through a simple table lookup. To our knowledge, no prior work has examined the effectiveness of non-contextual BERT embeddings. We investigate how these embeddings, extracted from the model's first layer, compare to FastText embeddings for tasks such as news classification, sentiment analysis, and hate speech detection in Marathi.

Additionally, past comparisons often used BERT's 768-dimensional embeddings against FastText's 300-dimensional ones, which is unfair since higher dimensions naturally provide better feature extraction. To address this, we ensure a fair comparison by reducing the BERT embeddings to 300 dimensions.

\begin{figure}
    \centering
    \includegraphics[width=1\linewidth]{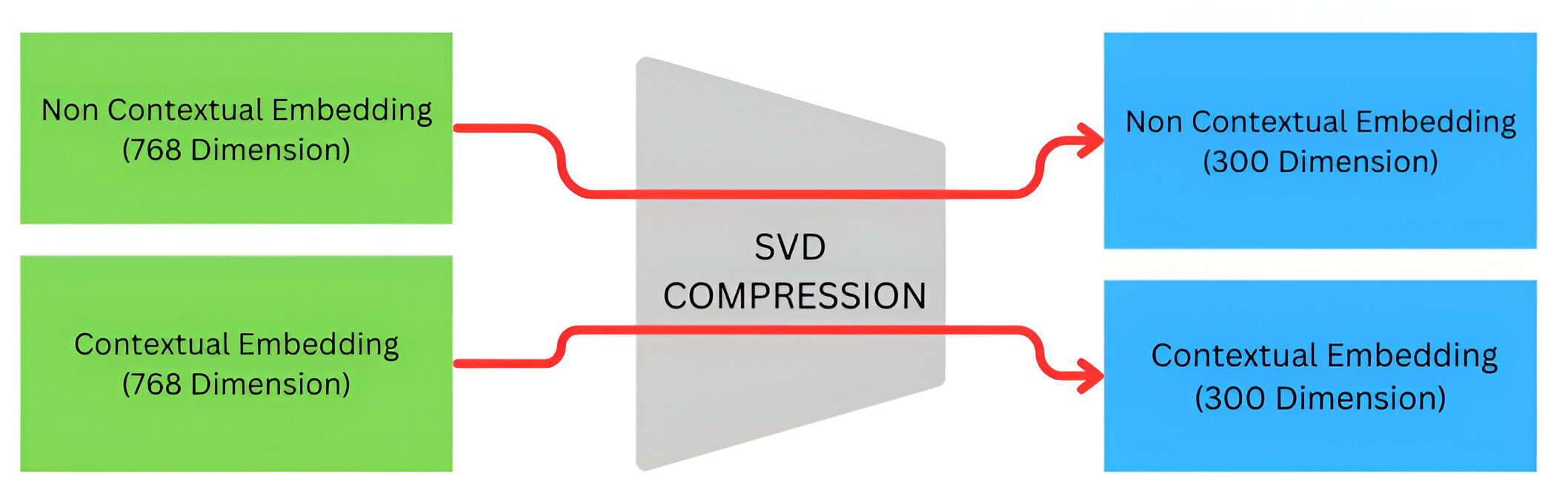}
    \caption{SVD compression of BERT embeddings}
    \label{fig:SVD_Arch}
\end{figure}

This paper focuses on utilizing FastText and non-contextual BERT for the Marathi language for the following tasks: Sentiment Classification, 2-Class and 4-Class Hate Speech Detection, and News Article Classification for headlines, long paragraphs, and long documents. We construct a comprehensive analysis of FastText embeddings, IndicFT \cite{kakwani-etal-2020-indicnlpsuite} and MahaFT \cite{joshi-2022-l3cube} embeddings, and BERT embeddings, including muril-base-cased \cite{khanuja2021muril} and marathi-bert-v2 \cite{joshi-2022-l3cube}. To enhance the comparison, we replicate the experiments using widely utilized contextual BERT embeddings. We also evaluate the impact of compression on both contextual and non-contextual BERT-based embeddings. Our analysis shows that non-contextual BERT embeddings generally perform better than FastText in most tasks. Furthermore, contextual BERT embeddings consistently outperform FastText across all evaluated tasks. However, compressing non-contextual embeddings reduces their performance, making FastText more effective than compressed non-contextual BERT.


The key contributions of this work are as follows: 
\begin{itemize}[leftmargin=*, label=\textbullet]
    \item We conduct a detailed study comparing non-contextual BERT embeddings and FastText embeddings for Marathi, a low-resource language. The evaluation covers multiple classification tasks, including sentiment analysis, news classification, and hate speech detection.
    
    \item To ensure a fair comparison, we compress BERT embeddings from 768 to 300 dimensions using Singular Value Decomposition (SVD). This allows us to analyze how dimensionality reduction impacts BERT's performance compared to its uncompressed version and FastText.
    
    \item We explore the differences between contextual and non-contextual BERT embeddings, examining their impact on classification performance in low-resource settings.
\end{itemize}

The paper is organized as follows: Section 2 provides a concise review of prior research on FastText and BERT. Section 3 includes the datasets and model embeddings that are utilized for the experiments. Section 4 presents the methodology used. Section 5 presents the results and key insights drawn from the findings along with a comparative analysis of FastText embeddings and BERT. In Section 6, we analyze our results and explain the reasons behind them. In Section 7, we conclude our discussion.

\section{Literature Review}

The existing literature emphasizes the superiority of contextual BERT embeddings over traditional word embedding techniques like Word2Vec \cite{mikolov2013efficientestimationwordrepresentations} , GloVe \cite{pennington2014glove}, and FastText across various natural language processing (NLP) tasks. For instance, \citet{khaled2023arabic} compare four popular pre-trained word embeddings—Word2Vec (via Aravec \cite{article}), GloVe, FastText, and contextual BERT (via ARBERTv2)—on Arabic news datasets. They highlight BERT’s superior performance, achieving over 95\% accuracy due to its contextual interpretation. 

Similarly, \citet{inproceedings2} analyzes the performance of embeddings on the AG News dataset, which includes 120K instances across four classes. They conclude that contextual BERT outperforms other methods, achieving 90.88\% accuracy. FastText, Skip-Gram, CBOW, and GloVe achieve 86.91\%, 85.82\%, 86.15\%, and 80.86\%, respectively. 

While traditional embeddings perform reasonably well, the consistent dominance of contextual BERT in complex tasks is also noted in sentiment analysis. For instance, \citet{Xie2024ModelingSA} explores how combining BERT and FastText embeddings enhances sentiment analysis in education, demonstrating that BERT’s contextual understanding, along with FastText’s ability to handle out-of-vocabulary words, improves generalization over unseen text.

In the domain of toxic speech classification, \citet{DSa2020BERTAF} utilize both contextual BERT and FastText embeddings to classify toxic comments in English, with BERT embeddings outperforming FastText. This trend continues in hate speech detection, where \citet{Rajput2021HateSD} find that neural network classifiers using contextual BERT embeddings perform better than those with FastText embeddings alone, further supporting BERT's effectiveness.

Additionally, \citet{Chanda2021EfficacyOB} assess contextual BERT embeddings against traditional context-free methods (GloVe, Skip-Gram, and FastText) for disaster prediction, demonstrating BERT’s superior performance in combination with traditional machine learning and deep learning methods.

For low-resource languages (LRLs), \citet{Ahmed2024EnhancingDP} examine methods like traditional TF-IDF, contextual BERT, and FastText embeddings within a CNN-BiLSTM architecture for detecting depressive texts in Bangla. Their results show that BERT embeddings yield the highest F1 score (84\%), indicating their dominance over other methods. This suggests that BERT's efficacy extends even to LRLs.

In medical applications, \citet{khan2024integrating} proposes integrating contextual BERT embeddings with SVM for prostate cancer prediction. By incorporating both numerical data and contextual information from clinical text, they achieve 95\% accuracy, far outperforming the 86\% accuracy achieved with numerical data alone.

Moreover, \citet{Malik2021ToxicSD} uses both contextual BERT and FastText embeddings to preprocess a dataset of conversations from Twitter and Facebook. Applying various machine learning and deep learning algorithms, they find that CNN yields the best results, further demonstrating BERT's capabilities.

Finally, while \citet{Asudani2023ImpactOW} offers a comprehensive analysis of traditional word embeddings alongside more advanced techniques like ELMo and contextual BERT, providing insight into commonly used datasets and models for benchmarking, \citet{Umer2022ImpactOC} highlights the versatility of FastText in various domains, despite BERT's consistently superior performance.

We note that the reviewed literature highlights the consistent superiority of BERT embeddings across various NLP tasks and domains. However, most existing studies focus mainly on contextual BERT embeddings, but not on non-contextual embeddings.  Moreover, these studies predominantly address high-resource languages, leaving low-resource languages like Marathi largely unexplored. In particular, there is a lack of research assessing the effectiveness of non-contextual BERT embeddings for Marathi. Additionally, the impact of dimensionality leveling, i.e. the efficacy of BERT embedding compression, has not been explored.

\begin{table*}[hbt]
\centering
\small
\resizebox{\textwidth}{!}{%
\begin{tabular}{|c|c|c|c|c|c|c|c|}
\hline
\textbf{Type} & \textbf{Model} & \multicolumn{1}{c|}{\textbf{MahaSent}} & \multicolumn{2}{c|}{\textbf{MahaHate}} & \multicolumn{3}{c|}{\textbf{MahaNews}} \\ 
 &  & \href{https://github.com/l3cube-pune/MarathiNLP/tree/main/L3CubeMahaSent\%20Dataset}{\textbf{3-class}} & \href{https://github.com/l3cube-pune/MarathiNLP/tree/main/L3Cube-MahaHate/4-class}{\textbf{4-class}} & \href{https://github.com/l3cube-pune/MarathiNLP/tree/main/L3Cube-MahaHate/2-class}{\textbf{2-class}} & \href{https://github.com/l3cube-pune/MarathiNLP/tree/main/L3Cube-MahaNews/SHC}{\textbf{SHC}} & \href{https://github.com/l3cube-pune/MarathiNLP/tree/main/L3Cube-MahaNews/LDC}{\textbf{LDC}} & \href{https://github.com/l3cube-pune/MarathiNLP/tree/main/L3Cube-MahaNews/LPC}{\textbf{LPC}} \\ \hline

Contextual & \href{https://huggingface.co/l3cube-pune/marathi-bert-v2}{MahaBERT} & 82.27 & 66.8 & 85.57 & 89.83 & 93.87 & 87.78 \\
 & \href{https://huggingface.co/l3cube-pune/marathi-bert-v2}{MahaBERT (Compressed)} & 82.89 & 66.15 & 84.37 & 89.61 & 93.53 & 87.82 \\
 & \href{https://huggingface.co/google/muril-base-cased}{Muril} & 81.64 & 64.55 & 84.00 & 89.54 & 93.64 & 87.33 \\
\textbf & \href{https://huggingface.co/google/muril-base-cased}{Muril (Compressed)} & 81.91 & 63.2 & 83.36 & 88.38 & 93.48 & 87.45 \\
\hline
 FastText & \href{https://storage.googleapis.com/ai4bharat-public-indic-nlp-corpora/data/monolingual/indicnlp_v1/sentence/mr.txt.gz}{IndicFT} & 76.4 & 58.25 & 80.13 & 85.57 & 92.15 & 79.19 \\
 & \href{https://huggingface.co/l3cube-pune/marathi-fast-text-embedding}{MahaFT} & 78.62 & 62.75 & 81.76& 85.89 & 92.62 & 80.32 \\
 \hline
 Non-Contextual & \href{https://huggingface.co/l3cube-pune/marathi-bert-v2}{MahaBERT} & 77.56 & 66.5 & 82.64 & 86.45 & 91.69 & 81.76 \\
& \href{https://huggingface.co/l3cube-pune/marathi-bert-v2}{MahaBERT (Compressed)} & 76.31 & 63.9 & 81.57 & 83.85 & 91.25 & 80.08 \\
 & \href{https://huggingface.co/google/muril-base-cased}{Muril} & 76.58 & 65.77 & 81.79& 85.95 & 91.61 & 81.36 \\
 & \href{https://huggingface.co/google/muril-base-cased}{Muril (Compressed)} & 75.16 & 63.25 & 81.44 & 82.72 & 90.39 & 79.00 \\ \hline

\end{tabular}%
}
\caption{Performance of model embeddings on MahaSent, MahaHate, and MahaNews datasets using Multiple Logistic Regression.}
\label{tab:result}

\end{table*}
\begin{figure*}[h]
    \centering
    \includegraphics[width=1\linewidth]{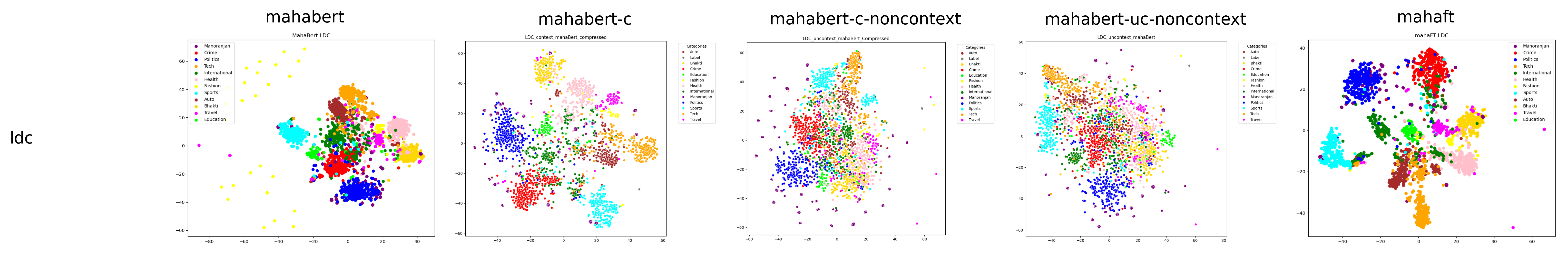}
    \caption{T-SNE Plot For BERT and FastText Embeddings 
    (c stands for compressed) .}
    \label{fig:TSNE Plot for LDC}
\end{figure*}

\begin{table*}[hbt]
    \centering
    \small
  \begin{tabular}{|l| l| l |c |c |c| c|}
    \hline
    \textbf{Dataset} & \textbf{Subdataset} & \textbf{Model} & \textbf{Avg} & \textbf{Variance} & \textbf{Std} & \textbf{Test} \\
    \hline
    MahaSent & 3 Class & MahaBERT & 76.56 & 0.39843 & 0.6312 & 78.01 \\
             &         & MahaBERT-Compressed & 74.42 & 0.8498 &\textbf{ 0.9218} & 75.51 \\
             &         & Muril & 75.53 & 0.75268 &\textbf{ 0.8676} & 76.53 \\
             &         & Muril-Compressed & 72.97 & 0.48963 & 0.6997 & 75.2 \\
             &         & MahaFT & 77.28 & 0.38282 & 0.6187 & 78.58 \\
    \hline
    MahaHate & 4 Class & MahaBERT & 64.92 & 0.25203 & 0.5020 & 66.1 \\
             &         & MahaBERT-Compressed & 62.77 & 0.53875 & 0.7340 & 64.1 \\
             &         & Muril & 63.51 & 0.35307 & 0.5942 & 65.15 \\
             &         & Muril-Compressed & 61.22 & 0.52378 & 0.7237 & 62.9 \\
             &         & MahaFT & 62.48 & 0.22608 & 0.4755 & 62.55 \\
    \hline
             & 2 Class & MahaBERT & 84.23 & 0.37633 & 0.6135 & 82.53 \\
             &         & MahaBERT-Compressed & 82.3 & 0.10312 & 0.3211 & 81.41 \\
             &         & Muril & 83.69 & 0.39397 & 0.6277 & 81.63 \\
             &         & Muril-Compressed & 81.67 & 0.20943 & 0.4576 & 81.41 \\
             &         & MahaFT & 83.75 & 0.52153 & 0.7222 & 82.61 \\
    \hline
    MahaNews & SHC & MahaBERT & 86.66 & 0.27687 & 0.5262 & 86.64 \\
             &     & MahaBERT-Compressed & 84.13 & 0.36002 & 0.6000 & 83.81 \\
             &     & Muril & 85.7 & 0.06973 & 0.2641 & 85.66 \\
             &     & Muril-Compressed & 82.89 & 0.11612 & 0.3408 & 82.01 \\
             &     & MahaFT & 87.25 & 0.17873 & 0.4228 & 85.97 \\
    \hline
             & LDC & MahaBERT & 92.47 & 0.32565 & 0.5707 & 91.69 \\
             &     & MahaBERT-Compressed & 91.41 & 0.01637 & 0.1279 & 91.57 \\
             &     & Muril & 92.03 & 0.19055 & 0.4365 & 91.69 \\
             &     & Muril-Compressed & 91.04 & 0.07753 & 0.2784 & 90.39 \\
             &     & MahaFT & 92.79 & 0.15667 & 0.3958 & 92.71 \\
    \hline
             & LPC & MahaBERT & 81.71 & 0.18503 & 0.4302 & 81.27 \\
             &     & MahaBERT-Compressed & 80.03 & 0.1779 & 0.4218 & 80.51 \\
             &     & Muril & 81.19 & 0.17597 & 0.4195 & 81.4 \\
             &     & Muril-Compressed & 78.82 & 0.14497 & 0.3807 & 79.11 \\
             &     & MahaFT & 80.15 & 1.25257 & \textbf{1.1192} & 80.32 \\
    \hline
\end{tabular}
        \caption{The values were obtained by performing 5-fold cross-validation on the training dataset for \textbf{Non-contextual embedding}. The \textbf{Avg}, \textbf{Variance} and \textbf{Std}  represent the average, variance and standard deviation respectively performance across the five test subsets (from training) of the 5-fold splits, while the \textbf{Test} column reflects the performance on the actual test dataset.}
\label{tab:kfold}
\end{table*}

\section{Datasets and Models Used}
In our research work, we used 3 Marathi datasets, \href{https://github.com/l3cube-pune/MarathiNLP/tree/main/L3CubeMahaSent\%20Dataset}{MahaSent}: A 3-class sentiment analysis dataset \cite{pingle2023l3cube}, \href{https://github.com/l3cube-pune/MarathiNLP/tree/main/L3Cube-MahaHate}{MahaHate}: A 2-class as well as a 4-class hate classification dataset \cite{patil2022l3cube} and \href{https://github.com/l3cube-pune/MarathiNLP/tree/main/L3Cube-MahaNews}{MahaNews}: A news categorization dataset with 12 classes \cite{mittal2023l3cube}.

We used two types of embeddings in our experiments: FastText and BERT embeddings.
For FastText, we utilized both \href{https://indicnlp.ai4bharat.org/FastText/#downloads}{IndicFT} \cite{kakwani-etal-2020-indicnlpsuite} and \href{https://huggingface.co/l3cube-pune/marathi-fast-text-embedding}{MahaFT} \cite{joshi-2022-l3cube} embeddings. This was because both models included a Marathi corpus as part of their training data. MahaFT, in particular, was specifically trained on a Marathi corpus, making it especially relevant for our experiments.
For BERT embeddings, we primarily used two BERT-based models: {\href{https://huggingface.co/l3cube-pune/marathi-bert-v2}{MahaBERT}} \cite{joshi-2022-l3cube} and \href{https://huggingface.co/google/muril-base-cased}{MuRIL} \cite{khanuja2021muril}.  Since both models were trained on Marathi data, we selected them to compare with the FastText embeddings.

\section{Methodology}

For each sentence, corresponding embeddings were generated and the corresponding categorical labels were encoded into numerical labels. The creation of BERT embeddings was done by first tokenizing the text using the BERT tokenizer, along with padding and truncation. The tokenized input was then passed to the model and the output of the last hidden layer of BERT was taken,
which was then averaged to get contextual embeddings for every sentence. Whereas for non-contextual embeddings, the output of the first embedding layer was used. Refer Fig \ref{fig:BERT_Arch} for the embedding extraction workflow diagram.

However, for FastText, which is a non-contextual embedding by default, the process was slightly different due to the lack of a predefined vocabulary. Unlike BERT, which employs a tokenizer capable of processing entire Marathi sentences, FastText necessitates the creation of a custom vocabulary. To achieve this, the training and validation datasets were concatenated and passed through a text vectorizer, which generated vectors for every word in the dataset. The vectorizer returned the vocabulary as a list of words in decreasing order of their frequency. The FastText model was then loaded using the FastText library, and for each word in the vocabulary, a word vector was retrieved to construct the embedding matrix. For each sentence, the text was split into individual words, and the corresponding embeddings were retrieved from the embedding matrix. These embeddings were then averaged to produce the final sentence embeddings.\\
Additionally, we experimented with compressed embeddings by reducing the dimensionality from 768 (the traditional BERT embedding dimension)
to 300. This compression was performed using
Singular Value Decomposition (SVD) to select the most relevant features, extracting the top 300 components for all the combinations of contextual as well as non-contextual for MahaBERT as well as Muril. Refer Fig \ref{fig:SVD_Arch} for compression flow diagram. \\ Feature scaling was also applied to the outputs. All embeddings were then passed to a multiple logistic regression(MLR) classifier for classification into target labels.

\subsection{Experimental Setup}
The experiments were conducted on Kaggle notebooks equipped with a P100 GPU accelerator, utilizing 16 GB of GPU memory, 20 GB of storage, and 32 GB of RAM. Accuracy was chosen as the evaluation metric, given the balanced nature of the datasets. For classification, the results obtained from the embeddings were mapped to final labels using a multinomial logistic regression model to maintain methodological simplicity.
To determine the validity of the results obtained, 5-fold cross-validation was performed for all tasks, and the results are presented in Table~\ref{tab:kfold}.

\subsection{Visualisation of Embeddings}

To visualize how BERT and FastText embedding can separate the classes, we plotted \href{https://scikit-learn.org/stable/modules/generated/sklearn.manifold.TSNE.html}{T-SNE} \cite{JMLR:v9:vandermaaten08a} graphs for the \href{https://github.com/l3cube-pune/MarathiNLP/tree/main/L3Cube-MahaNews/LDC}{LDC} dataset. We have 5 plots, with 4 plots for MahaBERT and 1 for MahaFT. Refer figure \ref{fig:TSNE Plot for LDC}.

\section{Results}

Table \ref{tab:result} presents the results for various embeddings, including MahaBERT, MuRIL, MahaFT, and IndicFT, across multiple datasets and tasks. It includes both contextual and non-contextual embeddings, as well as the compressed variants of MahaBERT and MuRIL. 

In sections \ref{subsec: conVSfast} and \ref{subsec: nonconVSfast}, we have considered the uncompressed versions of Muril and MahaBERT.
Further, in section \ref{subsec: compressionEffect}, we specifically show the effect of compression on Muril and MahaBERT. 

\subsection{Contextual vs FastText}
\label{subsec: conVSfast}
From Table \ref{tab:result}, we observe the following trend when comparing contextual embeddings with FastText embeddings: MahaBERT > MuRIL > MahaFT > IndicFT.

\subsection{Non-Contextual  vs FastText}
\label{subsec: nonconVSfast}
The trend of comparing non-contextual embeddings with FastText typically follows this order: MahaBERT > MuRIL > MahaFT > IndicFT. However, there are exceptions for the MahaSent and LDC datasets.

For these two datasets, FastText tends to perform slightly better. However, the difference is minimal, so we refer to Table \ref{tab:kfold} to determine whether this deviation is significant or simply random noise. We observe a high variance in MahaSent, suggesting that its deviation from the usual trend when comparing non-contextual embeddings with FastText may be attributed to noise and is unlikely to be significant.

In contrast, the LDC dataset also deviates from the trend but exhibits relatively low variance. As a result, for the LDC dataset, the performance trend when comparing non-contextual embeddings with FastText is as follows: MahaFT > IndicFT > MahaBERT > MuRIL.

\subsection{Effect of Compression}
\label{subsec: compressionEffect}

From Table \ref{tab:result}, it can be inferred that compression negatively impacts non-contextual embeddings, as uncompressed versions generally perform better. This is evident from MahaFT outperforming the compressed non-contextual MahaBERT embeddings in all datasets except MahaHate-4c, suggesting that compression lowers the performance of non-contextual BERT embeddings.

However, the effect of compression on contextual embeddings varies across datasets, making it challenging to derive a consistent conclusion.




\section{Inference}
In this section, we explain why the non-contextual MahaBERT embeddings outperform FastText (MahaFT and IndicFT) embeddings. Both MahaBERT and MahaFT embeddings have been trained on the same corpus of 752 million tokens \citet{joshi-2022-l3cube}. The superior performance of non-contextual MahaBERT embeddings can be attributed to its larger embedding size, training data size, and contextual training objective. Specifically, the embedding size for Marathi-BERT-v2 is 152M (197,285 × 768), compared to MahaFT, which is 132M (439,247 × 300).

IndicFT performs worse than MahaFT, likely due to its smaller dataset size of 551 million tokens \cite{kakwani-etal-2020-indicnlpsuite}. On the other hand, contextual BERT achieves better results because its hidden layers are effectively utilized.

Additionally, we observe a negative impact when compressing MahaBERT non-contextual embeddings. Reducing the embedding size from 152M (197,285 × 768) to 59M (197,285 × 300) leads to a decrease in performance, likely due to the loss of representational capacity.

\section{Conclusion}
In our research, we analyzed the effectiveness of various BERT and FastText-based embeddings on three key NLP tasks for Marathi: news classification, hate speech classification, and sentiment classification focusing primarily on non-contextualised BERT embeddings. 

Our results show that contextual BERT embeddings perform better than non-contextual ones, including both non-contextual BERT embeddings and FastText. Among non-contextual embeddings, BERT generally outperforms FastText in most tasks. However, when non-contextual BERT embeddings are compressed, their performance drops, and FastText performs better than compressed non-contextual BERT.


 \section*{Acknowledgement}
This work was carried out under the mentorship of L3Cube, Pune. We would like to express our gratitude towards our mentor, for his continuous support and encouragement. This work is a part of the L3Cube-MahaNLP project \cite{joshi2022l3cube}.

\bibliography{main}
\begin{figure*}[ht]
    \centering
    \includegraphics[width=1\linewidth]{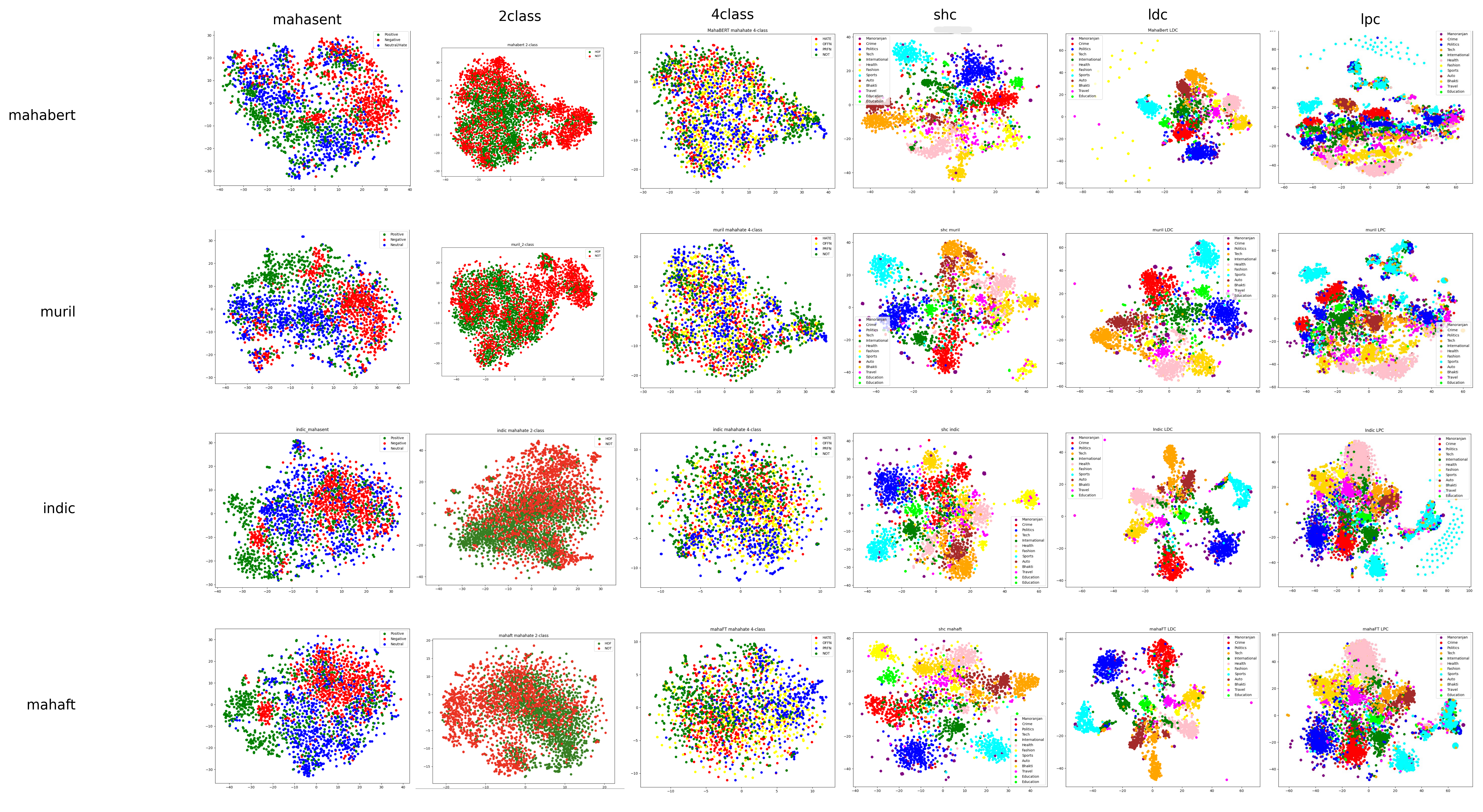}
    \caption{T-SNE Visualisation}
    \label{fig:TSNE Visulaisation across all datasets}
\end{figure*}
\appendix
\newpage

\section{Appendix}



\subsection{T-SNE Visualisations}
Refer to Fig. \ref{fig:TSNE Visulaisation across all datasets} for the t-SNE visualization of all datasets.

\end{document}